%% file: main.tex
\documentclass[10pt,twocolumn,letterpaper]{article}

\usepackage[pagenumbers]{iccv} 


\definecolor{iccvblue}{rgb}{0.21,0.49,0.74}
\usepackage[pagebackref,breaklinks,colorlinks,allcolors=iccvblue]{hyperref}

\def\paperID{*****} 
\def\confName{ICCV}
\def\confYear{2025}

\title{SAMSON: 3rd Place Solution of LSVOS 2025 VOS Challenge}

\author{
Yujie Xie\textsuperscript{1}\\
{\tt\small xyj@pixcakeai.com}
\and
Hongyang Zhang\textsuperscript{1,2}\\
{\tt\small hongyangzhang1@link.cuhk.edu.cn}
\and
Zhihui Liu\textsuperscript{1}\\
{\tt\small lzh@pixcakeai.com}
\and
Shihai Ruan\textsuperscript{1}\\
{\tt\small rsh@pixcakeai.com} \\
\textsuperscript{1}Truesight Research \\
\textsuperscript{2}School of Science and Engineering, The Chinese University of Hong Kong, Shenzhen
}

\begin{document}
\maketitle
\input{sec/0_abstract}    
\input{sec/1_intro}

\input{sec/2_related_work}
\input{sec/3_methods}
\input{sec/4_experiments}
\input{sec/5_conclusions}
{
    \small
    \bibliographystyle{ieeenat_fullname}
    \bibliography{main}
}

\end{document}

%% file: sec/0_abstract.tex
\begin{abstract}
Large-scale Video Object Segmentation (LSVOS) addresses the challenge of accurately tracking and segmenting objects in long video sequences, where difficulties stem from object reappearance, small-scale targets, heavy occlusions, and crowded scenes. Existing approaches predominantly adopt SAM2-based frameworks with various memory mechanisms for complex video mask generation. In this report, we proposed Segment Anything with Memory Strengthened Object Navigation (SAMSON), the \textbf{3rd place solution} in the MOSE track of ICCV 2025, which integrates the strengths of state-of-the-art VOS models into an effective paradigm. To handle visually similar instances and long-term object disappearance in MOSE, we incorporate a \textbf{long-term memory} module for reliable object re-identification. Additionly, we adopt \textbf {SAM2Long} as a post-processing strategy to reduce error accumulation and enhance segmentation stability in long video sequences. Our method achieved a final performance of 0.8427 in terms of $\mathcal{J}\&\mathcal{F}$ in the test-set leaderboard.
\end{abstract}

%% file: sec/1_intro.tex
\section{Introduction}
\label{sec:intro}

Video object segmentation (VOS) is a fundamental problem in computer vision, which aims to segment an arbitrary target throughout a video sequence, given a single annotated mask in the first frame \cite{r14, r15, r18}. 
While conventional VOS benchmarks are typically short video clips, long-term video object segmentation (LVOS) \cite{r16, r17, r19} extends this setting to much longer sequences, where substantial appearance changes, occlusions, and scene variations pose additional challenges.  The 7th LSVOS challenge tackles these difficulties through three tracks: Complex VOS, standard VOS, and Referring VOS, supported by the MOSEv2~\cite{r19}, MOSEv1~\cite{r17}, and MeViS\cite{r20}, datasets, respectively. Our method is primarily developed for Track 2, which is based on the MOSEv1 dataset.

The MOSEv1 dataset was introduced to advance video object segmentation (VOS) in complex scenes. Its successor, MOSEv2, presents a more challenging benchmark by intensifying scene complexity and introducing underrepresented factors, including adverse weather (rain, snow, fog), low-light conditions (nighttime, underwater), and multi-shot sequences. These enhancements aim to better approximate real-world scenarios and narrow the gap between existing VOS benchmarks and unconstrained environments. MeViS aims to segment and track target objects in videos based on natural language descriptions of their motions. This dataset involves handling diverse motion expressions that specify target objects within complex environments. It provides a benchmark for advancing language-guided video segmentation, where motion expressions serve as a primary cue to enhance object segmentation in challenging video scenes.

Recent progress in video object segmentation (VOS) has increasingly emphasized memory-based approaches due to their clear advantages over alternatives. Building on image-based SAM, SAM2~\cite{r1} introduces a memory module that extends its capability to VOS tasks and delivers notable improvements in segmentation performance. Nevertheless, its greedy segmentation strategy remains vulnerable to challenging scenarios involving frequent occlusions and object reappearances, while the fixed 8-frame memory restricts its effectiveness in long-term video analysis. To address these limitations, Segment Concept (SeC)~\cite{r4} introduces a concept-driven paradigm that shifts from feature matching to constructing and leveraging high-level object representations. By equipping SAM2 with an enhanced long-term memory module, SeC achieves significant gains in VOS performance.

In this work, we propose SAMSON, our 3rd place solution for the MOSEv1 challenge. Witnessing the enhanced memory module of SeC, we further fine-tune its grounding encoder through a two-stage training strategy on the MOSEv2 dataset. To mitigate the error propagation inherent in the original SAM2 framework, we incorporate SAM2Long at inference, thereby improving segmentation robustness. Our approach achieves a $\mathcal{J}\&\mathcal{F}$ of 0.8427, with $\mathcal{J}=0.8182$ and $\mathcal{F}=0.8671$ on the MOSEv1 track of ICCV 2025. Beyond this primary solution designed for the competition, we also conducted a series of exploratory experiments to address the fundamental trade-off between memory length and computational cost from a different perspective. These explorations focused on designing a new memory paradigm, including  enlarging the temporal perception field and refining memory update mechanisms. Although this experimental approach is still in its preliminary stages, it offers valuable insights for tackling extreme challenges like long-term occlusions, which we discuss in a later section.

%% file: sec/2_related_work.tex
\section{Related Work}
\label{sec:formatting}

\subsection{Video Object Segmentation}
Video Object Segmentation (VOS) tasks, including semi-supervised and unsupervised segmentation, have been primarily evaluated on datasets like DAVIS~\cite{r14} and YouTube-VOS~\cite{r15}. DAVIS offers high-quality, short-term video sequences with dense annotations, focusing on precise segmentation under occlusions and appearance changes. YouTube-VOS provides larger-scale, diverse videos, introducing challenges such as dynamic backgrounds and moderate-length tracking. However, their short-to-medium durations limit their ability to assess long-term temporal consistency required for real-world applications.

Recent datasets such as LVOS~\cite{r16} and MOSE~\cite{r17} address long-term VOS with extended sequences featuring challenges like prolonged occlusions, object reappearances, and crowded scenes. LVOS emphasizes diverse categories and long-term tracking across thousands of frames, testing consistency, while MOSE focuses on occlusions and motion blur in cluttered environments, posing significant challenges. The core task is to maintain precision in long-term videos by combining the strengths of LVOS and MOSE. More recently, MOSEv2~\cite{r19} advances multi-object video segmentation by scaling data volume and diversity, with over 5,200 videos and 1.2M annotated frames. It covers broader categories, denser interactions, and harder conditions such as severe occlusions and illumination changes, providing a comprehensive benchmark for evaluating robustness and generalization in video segmentation models.

\subsection{Memory based methods in VOS}
Memory-based frameworks have become an emerging paradigm in video object segmentation, leveraging stored temporal information to ensure consistency across frames. Key methods include XMem~\cite{r7}, which uses an Atkinson-Shiffrin memory model for long-term VOS, partitioning memory into short-term and long-term components. However, it accumulates errors under heavy occlusions due to pixel-level matching without distractor filtering. Learning Quality-aware Dynamic Memory~\cite{r8} updates memory by feature quality, enhancing robustness. RMem~\cite{r9} restricts memory banks for relevant representations, lowering overhead. Still, both face memory overload and distractor interference in long-term, crowded videos.

Recent integrations with foundation models like SAM 2~\cite{r1} have advanced the VOS. SAM2Long~\cite{r6} uses a training-free memory tree for bidirectional propagation, addressing long-term occlusions but increasing complexity. DAM4SAM~\cite{r5} employs distractor-aware memory for tracking, suppressing similar objects, yet struggles with pure segmentation in extended sequences. SAMURAI~\cite{r10} leverages motion-aware memory for zero-shot tracking, but may miss static objects, risking context loss. Recently, SeC~\cite{r4} utilizes Large Vision-Language Models (LVLMs) to progressively construct high-level, object-centric representations by integrating visual cues across frames, enabling robust semantic reasoning. During inference, SeC builds comprehensive semantic representations from processed frames for accurate segmentation of subsequent frames. Additionally, SeC dynamically balances LVLM-based semantic reasoning with enhanced feature matching, adapting computational efforts to scene complexity.

However, they still face the challengings: memory inefficiency and overload in long sequences (LVOS), handling uncertainty from occlusions and deformations (MOSE).

%% file: sec/3_methods.tex
\section{Methods}

\subsection{Overview}
Given a video sequence with $T$ frames $\{I_t\}_{t=1}^T$, the ground-truth mask $M_1$ of the target objects are provided in the first frame. The goal is to predict segmentation masks $\{M_t\}_{t=2}^T$ for the remaining frames through the segmentation model $f_{\theta}(\cdot)$.

\noindent\textbf{Image Encoder.} 
We adopt Hiera \cite{r3}, a hierarchical masked autoencoder, as the image encoder. Its multiscale architecture enables effective capture of both local details and long-range dependencies, providing robust representations for video segmentation. 

\noindent\textbf{Mask Encoder.} 
The mask encoder in SAM2 encodes segmentation masks by first embedding the input mask through a convolutional module, which projects it into the feature space. This embedding is then element-wise combined with the corresponding frame features from the image encoder, followed by lightweight convolutional layers for feature fusion. 
During tracking, only initialization masks or predicted masks are used, while interactive inputs such as clicks or bounding boxes are excluded to ensure full automation. This design refines mask representations in a compact and efficient manner, enabling precise segmentation and seamless integration into the overall SAM2 pipeline.

\noindent\textbf{Memory Bank.} 
The memory bank stores the initialization frame with its ground-truth mask and the six most recent frames with predicted masks. 
Temporal encodings are applied to recent frames to preserve ordering, while the initialization frame remains unencoded to serve as a target prior. 

\noindent\textbf{Mask Decoder.} 
Current-frame features attend to memory frames to obtain memory-conditioned representations, which are decoded into three candidate masks with IoU scores. 
The mask with the highest score is selected as output, and the memory is updated in a first-in-first-out manner, with the initialization frame permanently retained.

\noindent\textbf{Optimization.}  
The training objective of the proposed method combines complementary losses for pixel-level accuracy, region alignment, overlap quality, and mask-score regression. Concretely, we use a binary cross-entropy (BCE) loss for pixel-wise foreground/background classification, an IoU loss for region-level alignment, a Dice loss to mitigate class imbalance, and a Mask loss to supervise the decoder's predicted mask quality scores.

\begin{equation}
\mathcal{L} = \lambda_{1}\mathcal{L}_{\text{BCE}} + 
              \lambda_{2}\mathcal{L}_{\text{IoU}} + 
              \lambda_{3}\mathcal{L}_{\text{Dice}} +
              \lambda_{4}\mathcal{L}_{\text{Mask}},
\end{equation}

\noindent where the $\mathcal{L}_{\text{Mask}}$ term is defined as:
\begin{equation}
\mathcal{L}_{\text{Mask}} = \frac{1}{K}\sum_{k=1}^{K} \ell\!\big(\hat{s}_k,\, s_k\big), \quad
s_k = \mathrm{IoU}(\hat{M}_k, M_{\text{gt}}),
\end{equation}
with $\hat{s}_k$ the decoder's predicted IoU for candidate mask $\hat{M}_k$, $s_k$ the ground-truth IoU computed against $M_{\text{gt}}$, $K$ the number of candidates per frame, and $\ell(\cdot,\cdot)$ a regression loss (e.g. Smooth-$L_1$ or MSE). The weights $\lambda_{1..4}$ balance the terms.

\begin{figure}[t!]
\centering
\includegraphics[width=2.8in,height=3.8in]{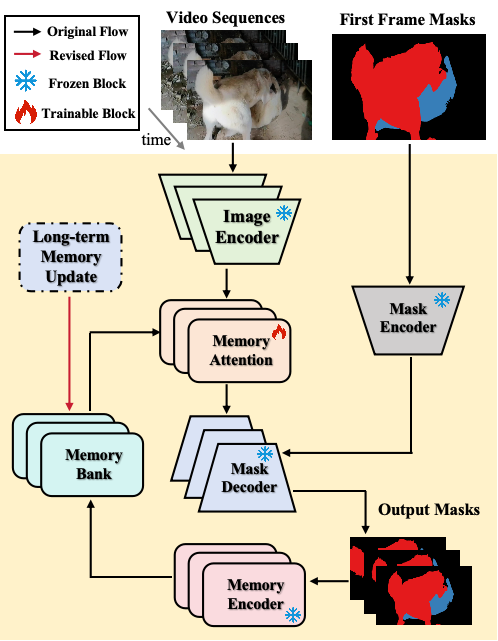}
\caption{Overview of the proposed second stage of training pipeline, where only the memory attention module is fine-tuned during this process.}
\label{fig:train}
\end{figure}

Since the SeC framework adaptively balances LVLM-based semantic reasoning with feature matching and dynamically allocates computation according to scene complexity, and given its superior empirical performance over state-of-the-art methods such as SAM2 and its variants across multiple benchmarks, we adopt it as our baseline. The training framework for the second stage is illustrated in Figure.\ref{fig:train}.

\begin{figure*}[ht!]
	\centering
	\includegraphics[width=\textwidth]{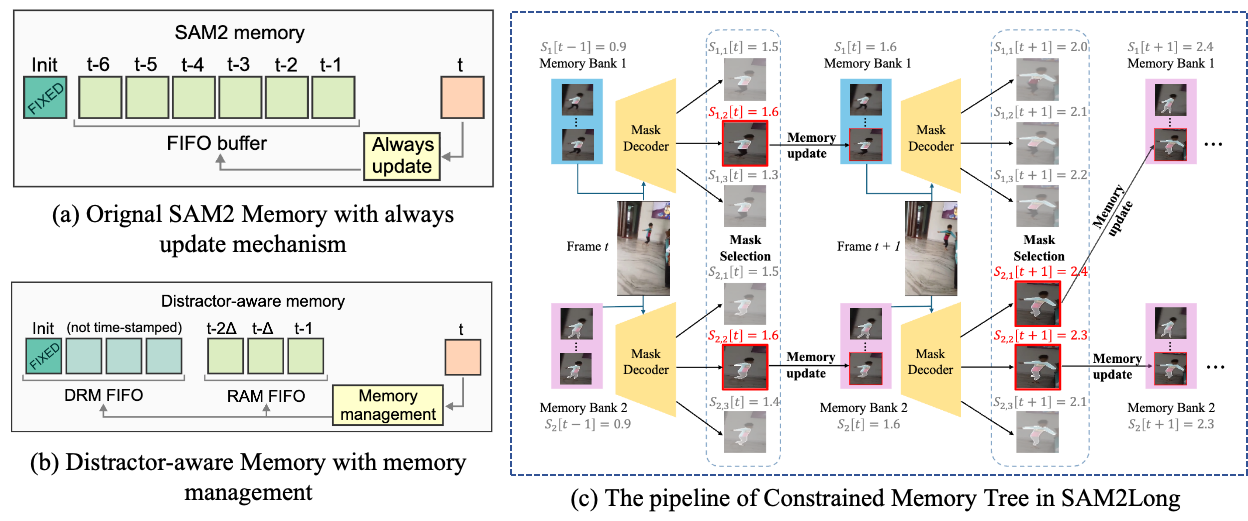}
	\caption{(a) Original design of the SAM2 memory mechanism and (b) proposed distractor-aware memory mechanism, both presented in \cite{r5}; (c) At each time step, multiple memory pathways are maintained, with the mask decoder generating candidate masks conditioned on memory banks. The pathway with the highest cumulative score is selected for propagation, adapted from \cite{r6}.}
	\label{fig:overview}
\end{figure*}

\subsection{Long-Term Memory Update for SAM}
We utilize the grounding encoder from SeC model and enhance the memory bank update mechanism by incorporating a distractor-aware memory module, drawing inspiration from DAM4SAM, to improve robustness and accuracy in video object segmentation. In the inference stage, SAM2Long~\cite{r6} is adopted for robust long-term video object segmentation, using a training-free memory tree to mitigate error accumulation and enable accurate tracking across extended sequences with occlusions.

\subsubsection{Sec Model}
Inspired by the Segment Concept (SeC) framework, we adopt its progressive concept-grounding encoder to construct high-level, object-centric representations for video object segmentation. SeC model is trained by the strategy as below:

\noindent\textbf{Concept Guidance with LVLM.}  
To strengthen concept-level reasoning, a sparse keyframe bank is maintained and updated during tracking. 
It retains the initialization frame and a few representative keyframes to ensure semantic diversity. 
LVLM encodes this compact set, with a special \texttt{<SEG>} token extracting object-level concept guidance. 

\noindent\textbf{Scene-Adaptive Activation.}  
To avoid redundancy, a scene-adaptive strategy applies concept guidance only when notable scene changes occur; otherwise, lightweight pixel-level matching is used. 
When activated, the LVLM-derived concept vector is fused with current frame features via cross-attention, enriching memory-enhanced representations. 
This balances semantic priors with fine-grained visual cues, ensuring robust segmentation across challenging scenarios.

\subsubsection{Distractor-Aware Memory Strategy}
To address long-term dependencies, we scale up the memory module inspired by the design of a distractor-aware memory (DAM). Figure~\ref{fig:overview} illustrates the memory management mechanisms in video object segmentation. Figure.\ref{fig:overview} (a) depicts the original SAM2 memory system with an always-update mechanism, featuring a FIFO buffer that processes frames from $t=6$ to $t=1$, with the initial frame fixed and the most recent frame always updated. Figure.\ref{fig:overview} (b) shows the distractor-aware memory management, incorporating a DFM FIFO for fixed initial frames (non-time-stamped), an RAM FIFO for dynamic frames from $t=2$ to $t=1$, and an integrated memory management module for enhanced robustness against distractors. The target of it mainly focused on tracking design, our memory stores an expanded set of temporal features \( \{F_t, C_t\}_{t=1}^T \), with a capacity increased by a factor of \( k \) (e.g., \( k=5 \)) relative to DAM. This larger memory retains detailed object information, critical for handling reappearances after prolonged occlusions. The distractor-aware mechanism computes a similarity score to filter irrelevant objects:
\begin{equation}
S_t = \text{Sim}(C_t, M_t), \quad M_t = \{F_i, C_i \mid i \in [1, t-1]\},
\end{equation}
where \( \text{Sim}(\cdot,\cdot) \) is a cosine similarity function, and \( M_t \) is the memory bank. Low-scoring distractors are suppressed, ensuring focus on the target object. The distractor-aware mechanism, adapted from DAM4SAM, enhances accuracy by mitigating interference from similar objects.

\subsection{SAM2Long for inference}
During inference, we further introduce SAM2Long to improve robustness without introducing additional training costs. The method adopts a constrained tree memory structure with uncertainty handling. 

The detailed information of constrained tree memory is illustrated in Figure.\ref{fig:overview} (c). Formally, given a set of memory nodes \(\{m_i\}_{i=1}^N\), each associated with an uncertainty score \(\sigma_i\), the aggregated memory feature at time step \(t\) is computed as:
\begin{equation}
\hat{M}_t = \sum_{i=1}^{N} w_i \cdot m_i, \quad 
w_i = \frac{\exp\left(-\sigma_i\right)}{\sum_{j=1}^N \exp\left(-\sigma_j\right)},
\end{equation}
where the weights \(w_i\) are constrained by the tree hierarchy, ensuring that closer parent-child nodes in the memory tree receive consistent weighting. 
The uncertainty score \(\sigma_i\) is estimated from prediction confidence, allowing unreliable memory nodes to be down-weighted automatically. 

The ensemble mechanism then fuses the uncertainty-aware memory \(\hat{M}_t\) with the concept representation \(C_t\), yielding the final segmentation prediction:
\begin{equation}
\hat{Y}_t = f_{\text{dec}}(\hat{M}_t, C_t, I_t),
\end{equation}
where \(I_t\) denotes the current frame embedding and \(f_{\text{dec}}\) is the mask decoder. 

This design not only balances adaptability and stability, but also mitigates error accumulation by dynamically suppressing noisy or outdated memory entries. 
Consequently, SAM2Long achieves consistent tracking under long-term occlusion, re-appearance, and large-scale appearance variations, while maintaining high efficiency at test time. 




%% file: sec/4_experiments.tex
\section{Experiment}

\subsection{Implementation details}
\noindent\textbf{Training Details.} In this experiment, we adopt SeC’s ground-encoder as the baseline, given its effectiveness demonstrated across multiple benchmarks, and fine-tune its memory encoder using the MOSEv2 dataset, which contains 3,666 annotated videos. The network architecture is based on DAM4SAM, while the pretrained weights are inherited from SeC.
MOSEv2 is chosen as it subsumes MOSEv1 while offering greater diversity and more challenging scenarios, thereby enhancing model generalization. The training process consists of two stages. In the first stage, we fine-tune the entire model with 8-frame inputs only due to the GPU memory constraints. In the second stage, we frozed all components except for the memory attention module and extended the input to 24 frames to recover the long-term memory capacity of the designed structure. All experiments is conducted on 8 H800 GPUs with a batch size of 1 per GPU. Input images are resized to \(1024 \times 1024\), with data augmentation applied through RandomHorizontalFlip, RandomAffine and ColorJitter. In the training process, the model is optimized by AdamW for 40 epochs. Besides, the optimization hyperparameter of $\lambda_{4}$ is set to 15.0.

\noindent\textbf{Benchmarks.}
We evaluated our model against two rigorous benchmarks: MOSE v1 and LVOSv1. MOSE v1 offers 2,149 video sequences, totaling over 560K frames, with dense annotations for multi-object video segmentation. This benchmark addresses diverse challenges such as occlusion, rapid motion, and scale variations. LVOSv1 comprises 1,128 long-term videos, exceeding 400K frames, with many sequences over 1,000 frames. This dataset is ideal for assessing robustness to occlusion, target re-appearance, and temporal consistency in long-term video segmentation. The validation sets include 311 video clips from MOSE v1 and 50 from LVOSv1.

\noindent\textbf{Metrics.} We adopt standard evaluation metrics: region similarity ($\mathcal{J}$), contour accuracy ($\mathcal{F}$) and their average value ($\mathcal{J}\&\mathcal{F}$) on the two benchmarks.




\subsection{Performance Comparsion}
We present a series of ablation studies on the validation
subsets on the two benchmarks, utilizing SAM2-Large as the default model size following the setting in SeC.

\begin{table*}[t]
\centering
\caption{Performance Comparsion ($\%$) on MOSEv1 and LVOSv1 datasets. All baselines are trained on MOSEv2 dataset, except for zero-shot. $^{*}$ denotes the model with fine-tuning tricks.}
\setlength{\tabcolsep}{3.5mm} 
\label{tab:video_seg_results}
\begin{tabular}{cccc|ccc}
\hline
\textbf{Method} & \multicolumn{3}{c|}{\textbf{MOSEv1 val}} & \multicolumn{3}{c}{\textbf{LVOSv1 val}} \\
 & $\mathcal{J}\&\mathcal{F}$ & $\mathcal{J}$ & $\mathcal{F}$ & $\mathcal{J}\&\mathcal{F}$ & $\mathcal{J}$ & $\mathcal{F}$ \\
\hline
baseline (zero-shot) & 75.1 & 71.1 & 79.1 & 77.4 & 73.2 & 81.6 \\
baseline + DAM & 77.7 & 73.6 & 81.8 & 81.2 & 77.6 & 84.9 \\
baseline + DAM$^{*}$ & 78.1 & 73.8 & 82.3 & 81.7 & 76.9 & 86.5 \\
baseline + DAM$^{*}$ + SAM2Long & 77.9 & 73.7 & 82.1 & 83.6 & 78.8 & 88.5 \\
\hline
\end{tabular}
\label{ab}
\end{table*}

\noindent\textbf{Effectiveness of the proposed module.} First, we explore the effectiveness of the DAM module and fine-tuning tricks in the training stage, SAM2Long in the inference stage. As shown in Table~\ref{ab}, introducing DAM brings clear performance improvements over the zero-shot baseline, e.g., from 75.1 to 77.7 in $\mathcal{J\&F}$ on MOSEv1 and from 77.4 to 81.2 on LVOSv1. Further applying fine-tuning tricks provides additional gains, especially on LVOSv1, where $\mathcal{F}$ improves from 84.9 to 86.5. Finally, equipping the model with SAM2Long during inference achieves the best overall results, reaching 83.6 in $\mathcal{J\&F}$ on LVOSv1 and 77.9 on MOSEv1, which demonstrates the complementary benefits of DAM and SAM2Long.

\begin{table*}[t]
\centering
\caption{Ablation Study Performance ($\%$) of different fine-tuning tricks on MOSEv1.}
\setlength{\tabcolsep}{3.5mm} 
\label{tab:video_seg_results}
\begin{tabular}{c|ccc}
\hline
\textbf{Method} & \multicolumn{3}{c}{\textbf{MOSEv1 val}} \\
 & $\mathcal{J}\&\mathcal{F}$ & $\mathcal{J}$ & $\mathcal{F}$ \\
\hline
baseline + Full FT & 77.7 & 73.6 & 81.8 \\
baseline + Memory FT & 77.0 & 72.8 & 81.3 \\
baseline + two-stage FT & 77.9 & 73.7 & 82.1 \\
\hline
\end{tabular}
\label{ft}
\end{table*}

\noindent\textbf{Effectiveness of Fine-tuning tricks.} Furthermore, we compare the performance of the model with full fine-tuning (Full FT), the model only fine-tuned with Memory Attention (Memory FT), and the model with two-stage fine-tuning (two-stage FT). In Table~\ref{ft}, two-stage FT achieves the best performance with a $J\&F$ score of 77.9, slightly surpassing Full FT (77.7), while Memory FT lags behind at 77.0. This demonstrates that selectively freezing modules and progressively unfreezing them not only reduces computational cost but also recovers long-term memory capability, yielding more robust segmentation.

\subsection{Qualitative Results}
\begin{figure*}[ht!]
	\centering
	\includegraphics[width=\textwidth]{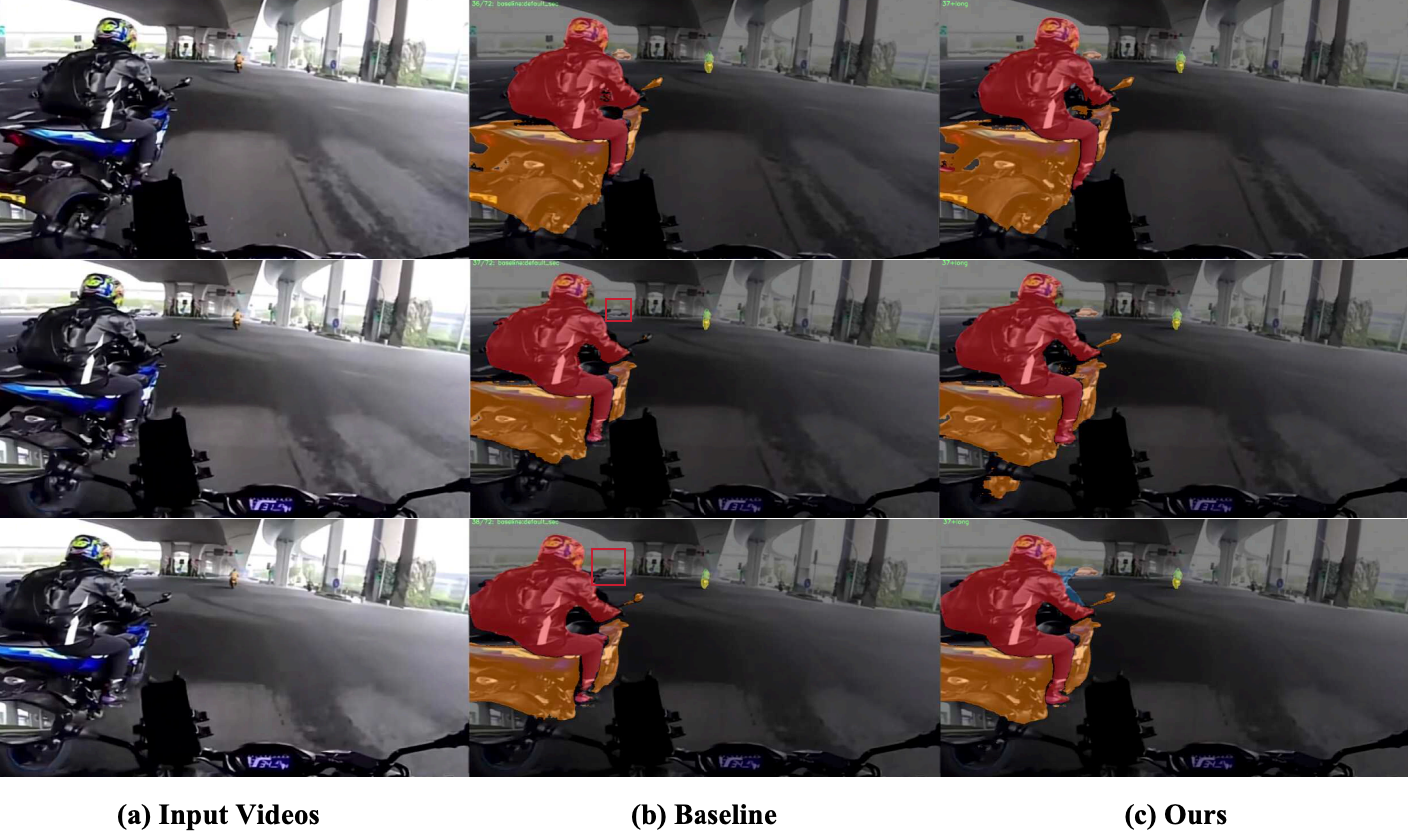}
	\caption{Qualitative comparison of video1 on MOSEv1 dataset. (a) Input video frames. (b) Results of the baseline method. (c) Results of our method. The red bounding box denote the failed examples in baseline.}
	\label{fig:ex1}
\end{figure*}

\begin{figure*}[ht!]
	\centering
	\includegraphics[width=\textwidth]{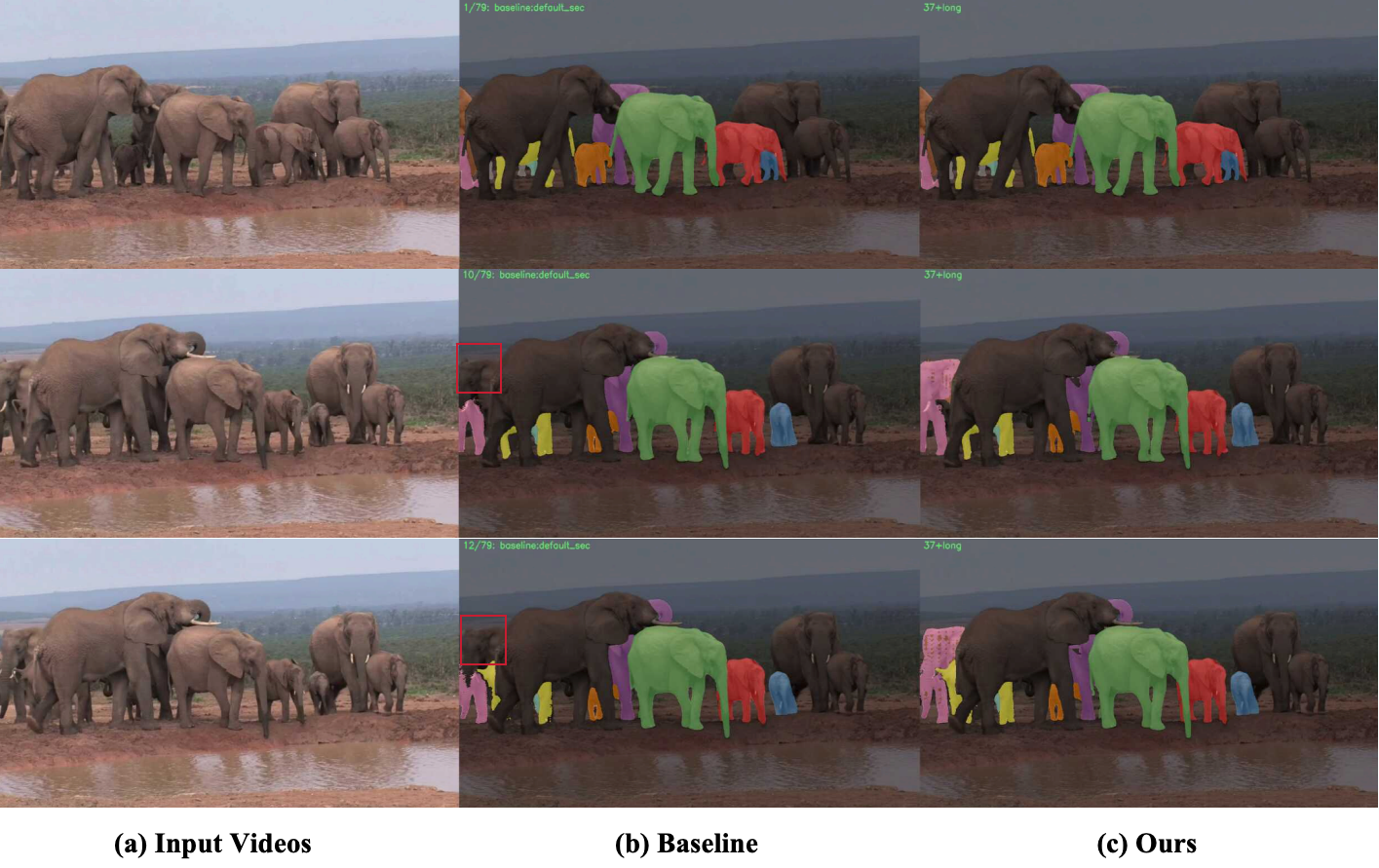}
	\caption{Qualitative comparison of video2 on MOSEv1 dataset. (a) Input video frames. (b) Results of the baseline method. (c) Results of our method. The red bounding boxs denote the failed examples in baseline.}
	\label{fig:ex2}
\end{figure*}

To evaluate the performance of the proposed method, we present the visualization results between Baseline and our proposed method in Figure \ref{fig:ex1} and Figure \ref{fig:ex2}, and the comparison clearly demonstrates that our method produces more accurate and consistent segmentation, especially in challenging scenarios with object occlusion, appearance changes, and cluttered backgrounds.

\section{Exploratory Study: An Enhanced Memory Paradigm}

\subsection{Motivation}

\begin{figure}[ht!]
	\centering
	\includegraphics[width=3.2in,height=2.2in]{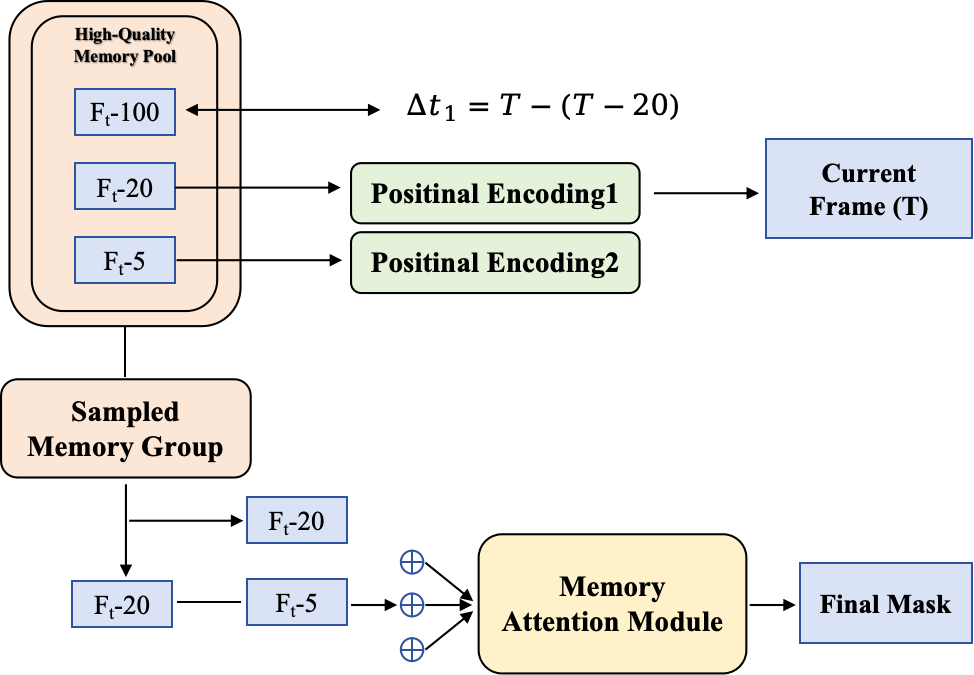}
	\caption{The pipeline of the enhanced Memory Module.}
	\label{fig:lt}
\end{figure}

We recognize a fundamental limitation in existing VOS models: fixed-length, first-in-first-out (FIFO) sliding window memory, as used in SAM2, is inherently constrained when handling long videos. This weakness becomes particularly evident during prolonged occlusions or significant appearance changes, where critical historical information is discarded, leading to tracking failures. To address the trade-off between memory length and computational cost, we conducted an exploratory study to design a more robust and flexible memory paradigm, shown in Figure \ref{fig:lt}. 

\subsection{Dynamic Multi-Group Memory Architecture}

Instead of simply enlarging memory capacity, our method fundamentally redesigns memory organization and utilization through four key mechanisms.

\noindent\textbf{Effective Frame Selection}. We establish an extensive "memory pool" to retain a large number of historical frames. To ensure memory quality, we filter frames using predefined IoU and object score thresholds before adding them to or sampling from the pool, guaranteeing that only high-confidence frames are used for segmentation. 

\noindent\textbf{Dynamic Positional Encoding}. We critically redesigned SAM2's positional encoding. Traditional methods use relative batch order, which fails to reflect true temporal separation. Our method dynamically calculates sinusoidal positional encoding based on the actual frame index difference between a memory frame and the current frame, normalized by a maximum temporal gap. This provides a more precise, physically intuitive representation of motion and temporal continuity, crucial for robust long-term tracking.
Following is the modified formula of the position encoding:
\begin{equation}
\text{PE}_{\text{orig}}(\text{pos}, d) = \sin\left( \frac{\text{pos}}{10000^{\frac{2d}{d_{\text{model}}}}} \right)
\end{equation}
\begin{equation}
\text{PE}_{\text{improved}}(t, p, d) = \sin\left( \frac{\min(|t-p|, 128)/128}{10000^{\frac{2d}{d_{\text{model}}}}} \right)
\end{equation}
where \(\text{PE}_{\text{orig}}(\text{pos}, d)\) represents the original position encoding in SAM2, \(\text{PE}_{\text{improved}}(t, p, d)\) represents our modified version. \(|t-p|\) is the frame index difference and is normalized by 128 (a preset maximum timestamp).

\noindent\textbf{Multi-Group Memory Processing}. At inference, we sample multiple memory groups from a high-quality pool: recent frames preserve short-term dynamics, while early frames maintain identity. This flexible strategy, supported by our training design, enables simultaneous exploitation of both local and global temporal context. 

\noindent\textbf{Weighted Fusion}. Pixel features from different memory groups are aggregated through weighted summation, enabling integration of target information across historical periods. This ensemble-like design improves robustness by reducing reliance on any single, potentially suboptimal memory. In the current implementation, weights are assigned uniformly, leaving room for future exploration of more adaptive strategies. 

\subsection{Preliminary Results and Analysis}

This approach showed strong potential in challenging scenarios. In qualitative tests with long-term object disappearance and reappearance, the model re-identified the target where baselines failed, highlighting the value of retaining longer, higher-quality history.

However, the approach also introduced new issues. Its overall $J\&F$ score on the MOSE dataset was lower than our final SAMSON submission, primarily due to a significant increase in false positives. We hypothesize that while the expanded memory provides richer historical context, the current filtering and simple fusion mechanisms are not yet sophisticated enough to consistently distinguish the target from visually similar distractors over long periods. This can cause the model to be misled by outdated or irrelevant memory cues, especially in crowded scenes.

%% file: sec/5_conclusions.tex
\section{Conclusion $\&$ Future work}

In this report, we presented SAMSON, our solution to the 7th LSVOS challenge, which ranked 3rd place in the MOSE track of ICCV 2025. By integrating a long-term memory module with SAM2 and adopting SAM2Long at inference, our method effectively mitigates the challenges of object reappearance, occlusions, and error accumulation in long video sequences. Leveraging a two-stage fine-tuning strategy on MOSEv2, SAMSON achieves a strong performance of $\mathcal{J}\&\mathcal{F}=0.8427$ on the MOSEv1 leaderboard, demonstrating the importance of memory-enhanced object navigation for large-scale video object segmentation. Furthermore, our exploratory study on an enhanced memory paradigm, while not yet achieving competitive overall accuracy, has shown clear promise for handling extreme long-term occlusions. The preliminary results suggest that a larger memory pool is a viable path forward. Future work will focus on developing more intelligent and adaptive memory sampling and weighting mechanisms. By incorporating attention or quality assessment modules to actively suppress irrelevant historical information, we aim to harness the full potential of an expanded memory pool without sacrificing precision, ultimately developing a more powerful and robust VOS model.